\documentclass[conference]{IEEEtran}
\IEEEoverridecommandlockouts
\usepackage{pifont} 
\usepackage{cite}
\usepackage{amsmath,amssymb,amsfonts}
  \usepackage{algorithmicx}
    \usepackage{algpseudocode}
    \usepackage{algorithm}
\usepackage{graphicx}
\usepackage{textcomp}
\usepackage{xcolor}
\usepackage{multirow}
\usepackage{enumitem}
\usepackage{comment}
\usepackage{listings}
\usepackage[left=0.63in,right=0.63in,top=0.72in,bottom=1.65in]{geometry}
\usepackage{tikz}
\usetikzlibrary{arrows.meta,positioning,fit,shapes.multipart,calc}
\def\BibTeX{{\rm B\kern-.05em{\sc i\kern-.025em b}\kern-.08em
    T\kern-.1667em\lower.7ex\hbox{E}\kern-.125emX}}

\begin{document}




\title{\huge Agentic AI-RAN: Enabling Intent-Driven, Explainable and Self-Evolving Open RAN Intelligence} 

\author{
\IEEEauthorblockN{
Zhizhou He\IEEEauthorrefmark{1}, \textit{Member, IEEE},
Yang Luo\IEEEauthorrefmark{1}, \textit{Member, IEEE},
Xinkai Liu\IEEEauthorrefmark{1}, \textit{Graduate Student Member, IEEE},\\
Mahdi Boloursaz Mashhadi\IEEEauthorrefmark{1}, \textit{Senior Member, IEEE}, 
Mohammad Shojafar\IEEEauthorrefmark{1}, \textit{Senior Member, IEEE},\\
Merouane Debbah\IEEEauthorrefmark{2}, \textit{Fellow, IEEE}, 
and Rahim Tafazolli\IEEEauthorrefmark{1}, \textit{Fellow, IEEE}
}
\IEEEauthorblockA{
\IEEEauthorrefmark{1}5G/6GIC, Institute for Communication Systems (ICS), University of Surrey, Guildford, UK\\
\{zhizhou.he, y.luo, xinkai.liu, m.boloursazmashhadi, m.shojafar, r.tafazolli\}@surrey.ac.uk
}
\IEEEauthorblockA{
\IEEEauthorrefmark{2}6G Research Center, Khalifa University, Abu Dhabi, UAE\\
merouane.debbah@ku.ac.ae
}
}



\maketitle

\begin{abstract}
The open radio access network (O-RAN) exposes rich control and telemetry interfaces across the non-real-time RAN intelligent controller (Non-RT RIC), near-real-time RIC (Near-RT RIC), and distributed units, but {also complicates the operation of} multi-tenant, multi-objective RANs in a safe and auditable manner. In parallel, agentic artificial intelligence (AI) systems with explicit planning, tool use, memory, and self-management offer a natural way to structure long-lived control loops. {This article studies how such agentic controllers can be brought into O-RAN. We contrast agentic controllers with conventional machine learning (ML)/reinforcement learning (RL) xApps and organize the O-RAN task landscape around three clusters: network slice life-cycle, radio resource management (RRM) closed loops, and cross-cutting security, privacy, and compliance.} We then introduce a {compact} set of agentic primitives {—Plan-Act-Observe-Reflect, skills as tool use, memory and evidence, and self-management gates—} and show, in a multi-cell O-RAN simulation, {that} they improve slice life-cycle and RRM performance {relative to} conventional baselines and ablations that remove individual primitives. {The framework achieves an average 8.83\% reduction in resource usage across three classic network slices.}
\end{abstract}

\begin{IEEEkeywords}
Open RAN, agentic AI, xApps, digital twin, LLMs.
\end{IEEEkeywords}

\section{Introduction}
\subsection{Motivation and Background}

The evolution toward the open radio access network (O-RAN) aims to enable vendor-neutral, flexible, and intelligent management of radio access infrastructures by disaggregating functions and exposing standardized controller interfaces \cite{polese2023understanding, polese2023empowering}. The O-RAN architecture employs a disaggregated RAN design with layered intelligence, utilizing non-real-time RAN intelligent controller (Non-RT RIC) and near-real-time RAN intelligent controller (Near-RT RIC) \cite{alam2025comprehensive}. Building on this foundation, industry and academia have demonstrated modular near-RT RIC applications (xApps) and non-RT RIC applications (rApps) for traffic steering and several other control functions \cite{ngo2024ran, sroka2024policy}.

Recent work pushes this vision toward artificial intelligence (AI)-native O-RAN and agentic architectures, where {learning-enabled controllers are integrated as native control entities across the Service Management and Orchestration (SMO), RICs, and O-Cloud, rather than as auxiliary post-hoc optimization modules} \cite{chatzimiltis2025agentic, dev2025advancedAI, brik2024explainable}. On the algorithmic side, reinforcement learning (RL) and multi-agent RL (MARL) have been adopted for RAN slicing and radio resource management (RRM) in RIC-centric designs, e.g., using double deep Q-networks (DQN) and deep RL agents to optimize spectrum allocation, power control, and slice-specific resource sharing \cite{he2025heterogeneous, chatzimiltis2025ai}. While these approaches outperform static heuristics, they typically assume fixed task definitions, single control objectives per agent, and extensive offline training. {Recent work has begun exploring large language model (LLM)-based agents for wireless networks \cite{xu2024large}, typically deploying a single LLM as a centralized decision-maker for tasks such as configuration generation or anomaly explanation.}

In parallel, the O-RAN Alliance and several vendors have developed digital-twin RAN (DT-RAN) platforms that mirror network states for training and validating AI models under realistic radio conditions \cite{he2025digital, he2025heterogeneous, elkael2025agentran}. However, existing solutions still treat learning controllers as monolithic black boxes, focused on narrow tasks with limited support for explicit goals, reusable skills, memory, or self-governance across O-RAN control layers. These limitations motivate the Agentic AI-RAN perspective developed in this work.

\subsection{Contribution}

This paper advocates an {Agentic AI-RAN} perspective, which models O-RAN control entities as goal-driven agents that operate across Non-RT, Near-RT, and real-time (RT) layers. Building on this perspective, we introduce a structured agentic framework and demonstrate its implications for O-RAN control through architectural design, methodological formalization, and simulation-based evaluation. The main contributions are summarized as follows:
\begin{itemize}
  \item {In contrast to existing LLM-for-RAN works that place a black-box LLM at a single time scale, each O-RAN control entity is treated as a goal-driven agent operating across Non-RT, Near-RT, and RT layers with explicit timing and budget awareness.}
  \item {Unlike monolithic RL/MARL controllers, the framework formalizes built-in guardrails (gating, schema validation, rollback) and multi-timescale memory as first-class primitives, with explicit architecture-level mitigation of LLM timeout, hallucination, and digital twin (DT) drift.}
  \item We formalize these agentic primitives and contrast them with traditional machine learning (ML)/RL and MARL controllers in O-RAN, highlighting how they reshape planning, execution, memory, and safeguards across the Non-RT, Near-RT, and RT domains.
  \item Through simulation, we show that combining planning, memory, gating, and active telemetry can improve service-level agreement (SLA) satisfaction, stability, and explainability while remaining interoperable with existing O-RAN specifications.
\end{itemize}

\section{Background and Preliminaries}
\subsection{O-RAN Architecture}
\begin{figure*}[t]
\centering
\includegraphics[width=0.8\textwidth]{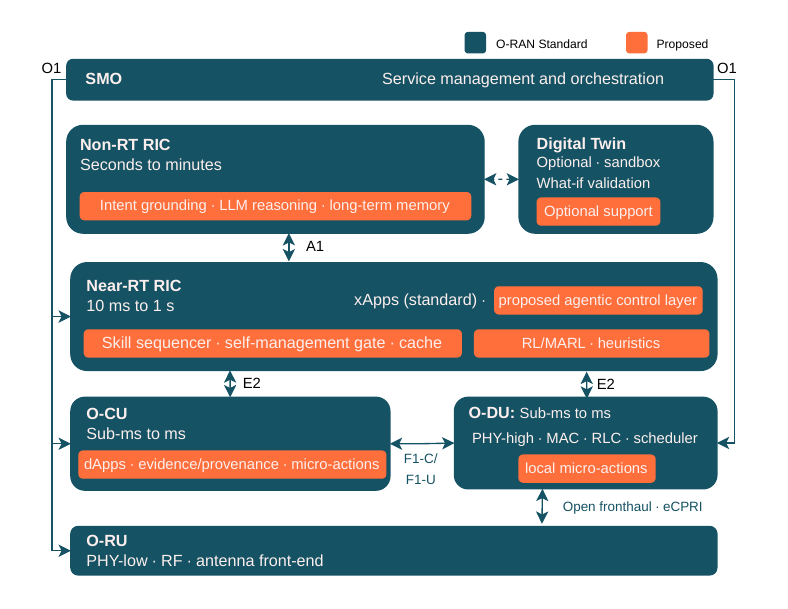} 
\caption{Proposed agentic AI-RAN architecture. Standard O-RAN entities and proposed agentic primitives are layered across the Non-RT RIC, Near-RT RIC, and O-CU/distributed application (dApp) tiers, communicating over the standard A1, E2, and Open Fronthaul interfaces.}
\label{fig:oran-archi}
\end{figure*}
The O-RAN architecture in Fig.~\ref{fig:oran-archi} adopts a disaggregated RAN design where radio and baseband functionalities are separated across the O-RAN Radio Unit (O-RU), O-RAN Distributed Unit (O-DU), and O-RAN Central Unit (O-CU). To enable programmable closed-loop control, O-RAN introduces the RAN Intelligent Controller (RIC) operating at multiple time scales: the Near-RT RIC supports near real-time control (10ms–1s), while the Non-RT RIC, integrated with the Service Management and Orchestration (SMO) framework, performs longer-timescale policy optimization and orchestration [14], [15]. Control functions are implemented via xApps (on the Near-RT RIC) and rApps (on the Non-RT RIC), interacting with the RAN through standardized interfaces such as E2 (control and telemetry) and A1 (policy guidance). The increasing number of concurrent xApps in dense deployments introduces challenges in coordination, conflict resolution, and lifecycle management, motivating more structured control paradigms.

\subsection{{AI for O-RAN: Approaches and Limitations}}
\label{subsec:ai-oran-limitations}

{Artificial intelligence has been widely adopted in O-RAN to enable adaptive control across multiple time scales. Supervised and deep learning models predict key performance indicators (KPIs) and recommend control actions from historical data, but lack adaptability under non-stationary conditions and offer limited interpretability. RL enables closed-loop xApps that adjust parameters such as handover thresholds and resource allocations, yet typically requires extensive training and exhibits instability in partially observable environments. MARL extends this to distributed control across RAN components, improving scalability at the cost of coordination overhead, non-stationarity, and inter-agent conflicts.}

{Beyond their individual weaknesses, these paradigms share four structural limitations relative to O-RAN deployment. First, they exhibit temporal misalignment with multi-layer O-RAN control loops, since most models are designed for a single time scale and hierarchical RL relies on pre-defined hierarchies that ignore cross-layer dependencies. Second, they encode objectives and constraints implicitly through reward functions or training data, which hampers constraint satisfaction, auditability, and safe operation under distribution shifts. Third, they generalize poorly to rare events and evolving traffic, often requiring retraining when conditions change. Fourth, they lack unified mechanisms for composing multi-step actions and enforcing global objectives across heterogeneous controllers. Table~I summarizes how these paradigms compare with the agentic perspective along axes of decision object, control cadence, safety guardrails, and O-RAN interface usage, motivating the structured control paradigm developed in the next section.}


\begin{table*}[t]
\centering
\caption{Paradigm comparison on key axes relevant to O\textendash RAN control.}
\label{tab:paradigm-fit}
\begin{tabular}{p{3.1cm}p{2.3cm}p{2.1cm}p{2.1cm}p{2.6cm}}
\hline
\textbf{Axis} & \textbf{Traditional ML} & \textbf{RL} & \textbf{MARL} & \textbf{Agentic AI} \\
\hline
Decision object & Static thresholds, predictions & Reactive policy & Interacting policies & Plans over skills with gating \\
Control cadence & Non\textendash RT (minutes\textendash hours) & Near\textendash RT feasible & Near\textendash RT with coordination cost & Non\textendash RT planning, Near\textendash RT sequencing, RT micro\textendash actions \\
Objective form & Single/multi\textendash target loss & Reward maximization & Team/competitive rewards & Multi\textendash objective with explicit budgets and constraints \\
Safety/guardrails & External & External & External + complex & Built\textendash in guardrails, rollback, evidence \\
Auditability & Low\textendash medium & Low & Low & High (decision provenance) \\
Data efficiency & High if labels exist & Medium\textendash low & Low (non\textendash stationarity) & Medium\textendash high via memory/retrieval-augmented generation (RAG) \\
Non\textendash stationarity & Sensitive & Sensitive & Challenging (credit assignment) & Mitigated via retrieval and small\textendash step control \\
O\textendash RAN interfaces & Indirect use of A1/E2 & Direct E2 actions & Coordinated E2 actions & Native tool\textendash use (A1/E2/E2 Service Model (E2SM)) \\
Typical placement & Non\textendash RT analytics & Near\textendash RT xApp & Near\textendash RT multi\textendash xApp & Cross\textendash layer (Non\textendash RT + Near\textendash RT + dApp) \\
Example tasks & KPI prediction, anomaly scoring & PRB split, HO tuning (steady) & Load balancing across cells/xApps & Slice admission/scale, rare\textendash event recovery, intent\textendash to\textendash policy \\
\hline
\end{tabular}
\end{table*}

\begin{figure*}[t]
\centering
\includegraphics[width=1.02\textwidth]{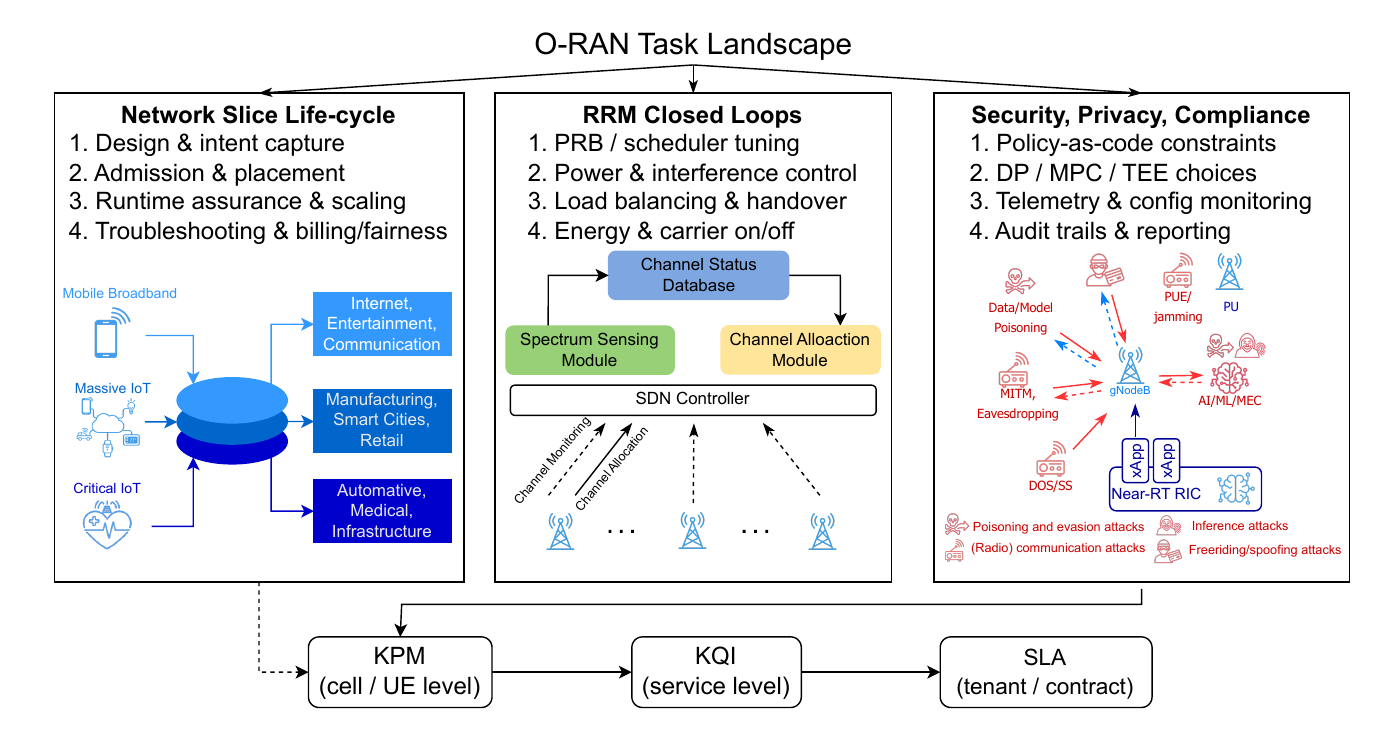} 
\caption{O-RAN task landscape with three clusters (network slice life-cycle, radio resource management closed loops, and security/privacy/compliance) and their relation to KPM, KQI, and SLA metrics.}
\label{fig:task-landscape}
\end{figure*}

\subsection{Security, Privacy, and Compliance}
\label{subsec:security-privacy}



Beyond slicing and RRM, O-RAN must satisfy security, privacy, and regulatory constraints in a disaggregated and multi-vendor environment. Existing deployments typically combine policy-driven control, secure interfaces, and runtime monitoring. At the Non-RT RIC/SMO layer, policy-as-code can encode constraints on data access, privacy preservation, and compliance, and disseminate them through A1 as operator intent. At the Near-RT layer, telemetry and constraint checks are used to detect violations related to security, privacy, or resource usage.

However, consistent enforcement remains difficult when performance optimization, compliance constraints, limited observability, and heterogeneous vendor implementations interact. This motivates an integrated control framework in which objectives, constraints, evidence, and safeguards are coordinated across O-RAN layers.

\section{Agentic AI for O-RAN: Problem Formulation and Design Rationale}
\begin{figure*}[t]
\centering
\includegraphics[width=1\textwidth]{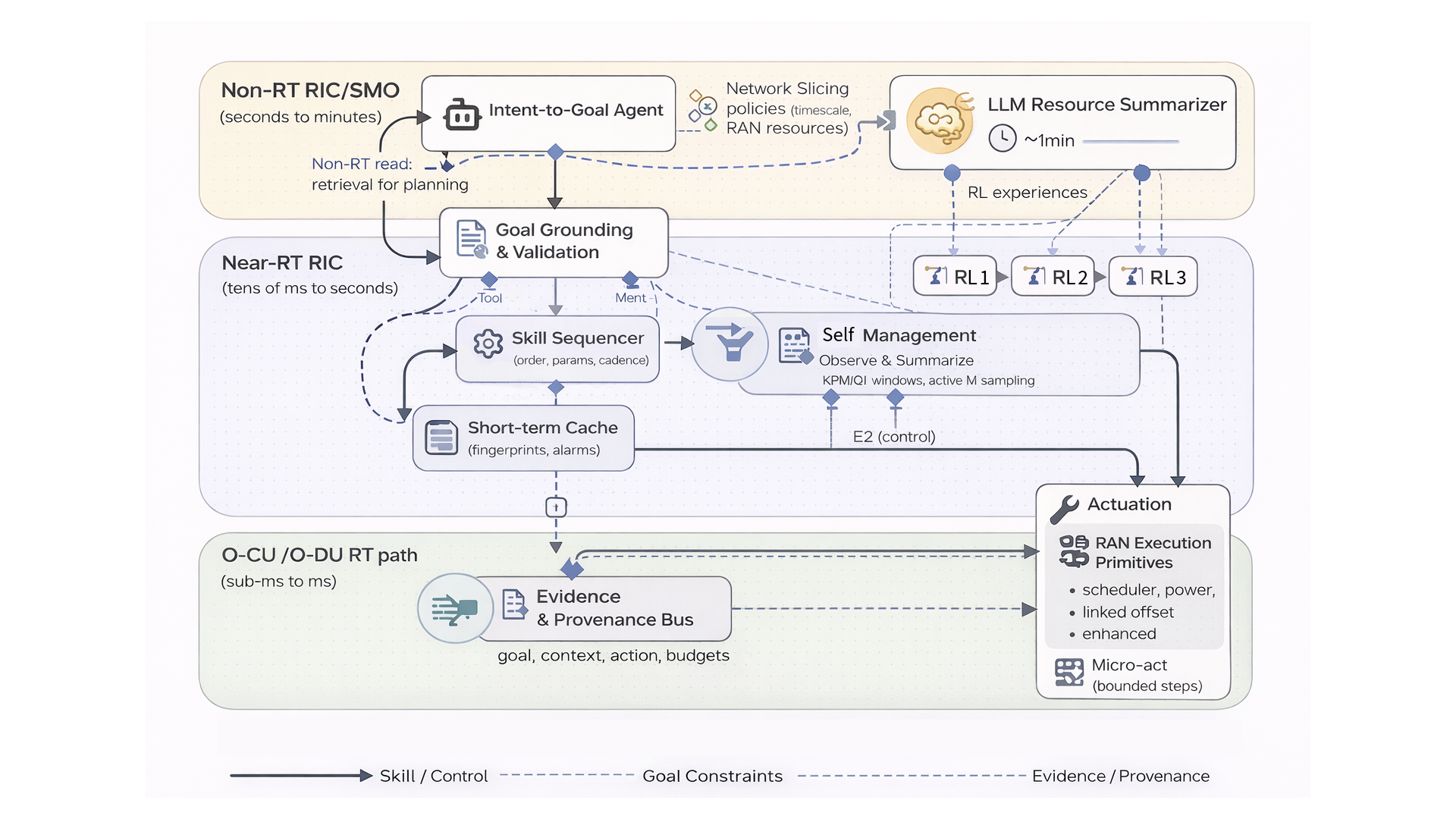} 
\caption{{Agentic primitives for O-RAN, aligned with control-loop cadences. The loop plans over skills, sequences and gates incremental commits, observes summarized Key Performance Measurement (KPM) and Key Quality Indicator (KQI) data, and records evidence. Memory spans short-term caches at Near-RT, episodic logs per decision, and long-term knowledge curated at Non-RT.}}
\label{fig:agentic-primitives-framework}
\end{figure*}

\subsection{System Overview}

Fig.~\ref{fig:oran-archi} illustrates the proposed agentic AI-RAN architecture. To respect the strict timing constraints of O-RAN control loops, higher-level reasoning is confined to the Non-RT RIC/SMO, while time-critical control remains at the Near-RT and real-time layers.
At the Non-RT layer, a high-capability decision module (e.g., an LLM or equivalent reasoning engine) operates at a coarse time scale (e.g., minute-level) to aggregate network-wide context, including KPM/KQI trends, resource utilization, and slice-level performance. Based on this global view, it derives intent-aligned policies such as slice-level resource allocation and controller selection, which are communicated to the Near-RT RIC via the A1 interface.
The Near-RT layer refines these policies under dynamic network conditions and executes control actions within strict latency budgets, while fine-grained actuation is performed at the O-CU/O-DU level. 

\subsection{What is “Agentic” in O-RAN?}

In the context of O-RAN, we define \emph{agentic control} as a structured decision-making paradigm in which control entities operate as goal-driven controllers that explicitly reason over objectives, constraints, and system context, rather than relying solely on fixed policies or reactive mappings. An agentic controller is characterized by three properties. First, it is \textbf{goal-oriented}: operator intents (e.g., latency, energy, or slice-level SLAs) are represented explicitly and guide decision-making beyond implicit reward signals or static configurations. Second, it is \textbf{deliberative}: instead of directly mapping observations to actions, the controller evaluates candidate actions or short action sequences under current system conditions. Third, it is \textbf{constrained}: decisions are subject to explicit limits on safety, resource usage, and control-loop timing, ensuring compatibility with O-RAN deployment requirements.

Importantly, agentic control does not replace existing control mechanisms in O-RAN but introduces an additional abstraction layer that coordinates them: heuristic controllers, ML predictors, and RL/MARL policies are treated as callable decision modules within a unified process. Agentic control thus focuses on \emph{how} decisions are structured and coordinated, rather than \emph{what} specific algorithms are used, distinguishing it from conventional ML, RL, and MARL approaches that optimize specific tasks at fixed time scales. {This distinction is one of abstraction level rather than of competing algorithm families. RL and MARL are \emph{content} optimizers: for a fixed task, objective, and time scale they learn \emph{what} action to take. Agentic control is an \emph{organizational} layer: it decides \emph{which} optimizer to invoke, \emph{when} and at what scope to commit its output, \emph{whether} that output is admissible under current budgets and constraints, and \emph{how} to compose and, if needed, roll back multi-step interventions across layers. The two are therefore complementary---an agentic layer is most effective precisely when it orchestrates strong content optimizers rather than substituting for them. Accordingly, the comparisons in this paper isolate the marginal value of the organizational layer while keeping the underlying content algorithms available, rather than pitting mutually exclusive alternatives against each other; the agentic column of Table~I should be read as a meta-controller that invokes the other paradigms as callable modules.}

\subsection{Decomposition Trade-offs}

A central premise of agentic control is the decomposition of decision-making across time scales, control layers, and functional modules. This decomposition improves modularity and scalability, but also introduces trade-offs that must be explicitly managed. Temporal decomposition separates long-horizon Non-RT reasoning from short-horizon Near-RT/RT execution, but may create mismatch between global objectives and local reactions. Functional decomposition enables reusable controllers for mobility, load balancing, slicing, and energy management, but these controllers may act on shared resources and generate conflicting decisions. Spatial decomposition improves scalability across cells, slices, or regions, but local controllers operate under partial observability and may deviate from system-level optimality. Finally, constraint-aware decomposition improves safety and compliance, but can restrict feasible actions and reduce optimization flexibility when constraints interact across layers.

These trade-offs show that decomposition cannot be treated as a purely structural design choice. Effective agentic control requires consistent propagation of objectives and constraints, mechanisms for resolving conflicts among decomposed controllers, and feedback pathways that account for delayed and cross-layer effects. The agentic formulation addresses these requirements by separating deliberation from execution while coordinating heterogeneous control modules across O-RAN layers.

\subsection{{Control Stability and Conflict Risks}}

{In distributed O-RAN environments, multiple control entities operating concurrently across overlapping resources can produce conflicting actions—e.g., when mobility, load balancing, and energy-saving functions independently adjust handover thresholds, transmission power, or resource allocation. Unresolved conflicts manifest as oscillatory behavior across consecutive decision cycles, increasing signaling overhead and degrading quality of service (QoS). In multi-agent settings, non-stationarity further compounds these effects, as each controller adapts to the actions of others. Formal stability guarantees are challenging in O-RAN due to multi-timescale control, partial observability, and stochastic environments; stability must instead be approximated through mechanisms that bound the rate, magnitude, and scope of control actions, introduce hysteresis or damping to avoid rapid reversals, prioritize or arbitrate among competing controllers, and incorporate delayed feedback into subsequent decisions. These factors are fundamental design constraints for any scalable O-RAN control paradigm.}

\subsection{Digital Twin Support}

A DT provides a simulation-based environment for planning, validation, and what-if analysis, enabling evaluation of candidate actions without impacting the live network. However, DT integration introduces practical challenges—synchronization drift, execution latency, and scalability limits—where discrepancies between simulated and real states can lead to inaccurate predictions. A DT provides a simulation-based environment for planning, validation, and what-if analysis, enabling evaluation of candidate actions without impacting the live network. However, DT integration introduces practical challenges---synchronization drift, execution latency, and scalability limits---where discrepancies between simulated and real states can lead to inaccurate predictions. {We therefore treat the DT as an \emph{architecturally optional but operationally active} module. Here ``optional'' denotes the absence of a hard dependency, not the absence of what-if capability: the DT is a self-contained module whose removal does not interrupt the control loop, so a DT outage or excessive synchronization drift degrades the system gracefully to live-only operation (Section~V-C) instead of stalling it. In the normal operating case the DT is deployed and runs as a lightweight parallel module with short per-query inference time, so it executes fast what-if rollouts \emph{online} rather than only during offline planning. In this online mode the DT acts as a buffer ahead of the live agents: it continuously pre-screens candidate skills and policies against rare or extreme scenarios (e.g., flash crowds, correlated cell outages, interference spikes) and pre-computes vetted fallbacks, so that when such conditions actually occur the Near-RT agents switch to a validated response instead of the exploratory excursions that would otherwise cause SLA and latency fluctuations. The what-if capability is thus retained during normal operation and is surrendered only transiently in the degraded mode, where conservative live-only safeguards take over.}

\section{Proposed Agentic Framework}
\subsection{Skills as Tool-Use}
\label{subsec:skills-tooluse}
{In our framework, we define a skill as a thin, verifiable wrapper around a controllable primitive that the agent invokes as a tool, structured with explicit preconditions, bounded actuation, expected effects, costs, and a compensating rollback. By binding skills to standard O\textendash RAN interfaces, the same abstraction applies consistently across the Non\textendash RT, Near\textendash RT, and RT layers. Skills are composable and incrementally committed: rather than executing indivisible plans, the agent runs short skill sequences, observes outcomes, and rolls back when risks or deviations exceed declared budgets. Local scoping and reversibility allow concurrent controllers to detect and order conflicts without centralized arbitration, supporting concise and auditable control sequences across heterogeneous deployments. For instance, an interference recovery scenario can be composed of three sequential skills: a power-cap skill on the offending cell, followed by physical resource block (PRB) reallocation to the affected slice, and a handover threshold adjustment to redistribute load. The No-Sequence ablation in Section VI cannot perform this composition and must address symptoms one at a time.}

\subsection{Memory and Evidence}
\label{subsec:memory-evidence}
{In our framework, the memory and evidence layer enables data-efficient and auditable agentic control. Memory spans multiple time scales: short-term state at Near\textendash RT for fast gating, episodic records of decisions and outcomes, and long-term knowledge at Non\textendash RT storing reusable policies and verified skill compositions. Retrieval informs plan synthesis and conservative execution under uncertainty, reducing reliance on costly retraining. Evidence is generated by design at each commit—decision-level records linking goals, compact context, selected actions, and observed outcomes—supporting audit and regulatory reporting without exposing raw subscriber data. Together, memory, retrieval, and evidence close the loop between fast execution and long-horizon learning across O\textendash RAN time scales.}

\subsection{Self-Management}
In our framework, self-management governs how an agent maintains safety and stability while pursuing goals under strict latency and budget constraints, keeping interventions predictable, reversible, and justifiable through evidence. {It is the mechanism that operationalizes the stability and conflict requirements of Section~III-D: rather than assuming formal stability guarantees, the gate explicitly bounds the \emph{rate}, \emph{magnitude}, and \emph{scope} of control actions, imposes hysteresis to suppress rapid reversals, arbitrates among competing controllers, and folds delayed feedback into subsequent decisions.} The core mechanism is a gate that evaluates each incremental commit against calibrated risk, resource budgets, and explanation consistency. Let $r_t$ denote predicted SLA-violation risk, $u_t$ an uncertainty score, $b_t$ current budget usage (CPU, E2 bytes, inference time), and $c_t$ an explanation-consistency score{; let $\delta_t$ denote the spatial scope of the commit (number of affected cells or PRB groups) and $\rho_t$ the recent commit rate within the current control window}. The agent advances only when all guards pass:
\begin{equation}
\begin{aligned}
\text{advance if: } & r_t \le \alpha,\ u_t \le \beta,\ b_t \le \mathcal{B},\ c_t \ge \tau{,\ \delta_t \le S,\ \rho_t \le R}; \\
\text{otherwise: } & \text{shrink step, wait, or rollback.}
\end{aligned}
\end{equation}
{The magnitude of each commit is bounded a priori by the per-skill actuation limits declared in Section~IV-A, so the gate jointly constrains rate ($R$), magnitude (per-skill caps), and scope ($S$)---the three quantities Section~III-D identifies as practical surrogates for stability.} When a guard fails, the step size $\gamma_t$ is reduced and a bounded observation window is inserted; repeated failures trigger rollback to a vetted baseline within RT/dApp deadlines. {To prevent the oscillatory reversals discussed in Section~III-D, a committed action on a given parameter cannot be reversed within a dwell window $W_d$ unless a hard safety guard fires, and a deadband suppresses commits triggered by sub-threshold KPM fluctuations. The guard thresholds are not static: under budget stress or detected non-stationarity they are tightened (smaller $\alpha,\beta$, larger $\tau$) and the step size is damped, then relaxed as conditions stabilize, yielding a self-regulating loop rather than a fixed admission test. Because cross-layer effects are delayed, a commit is not declared successful---and is therefore not promoted to long-term memory as a reusable composition---until its predicted KPM impact has been observed over a bounded credit window; pending effects are tracked in a per-commit ledger so that delayed degradations are attributed to the responsible action rather than to later, unrelated commits.} Concurrency is explicitly managed: the agent detects conflicts based on scope intersection and predicted KPM impact, enforcing {a deterministic priority ordering (safety/compliance $\succ$ SLA recovery $\succ$ efficiency) and scope locks that serialize commits on overlapping resources} (e.g., applying power caps before PRB reallocation in interference-limited regimes), and records resolutions as evidence to refine future arbitration. The same loop handles anomalies and budget stress through guard tightening, amplitude limiting, and fallback to safe baselines---with escalation to Non-RT recalibration when required---while each commit and rollback emits compact evidence for operator traceability. {In this way self-management also enforces the decomposition discipline of Section~III-C: it propagates Non-RT objectives downward as local risk and budget guards and re-aggregates local outcomes upward as evidence, keeping the decomposed controllers consistent with global objectives.}

\begin{figure*}[t]
\centering
\includegraphics[width=0.75\textwidth]{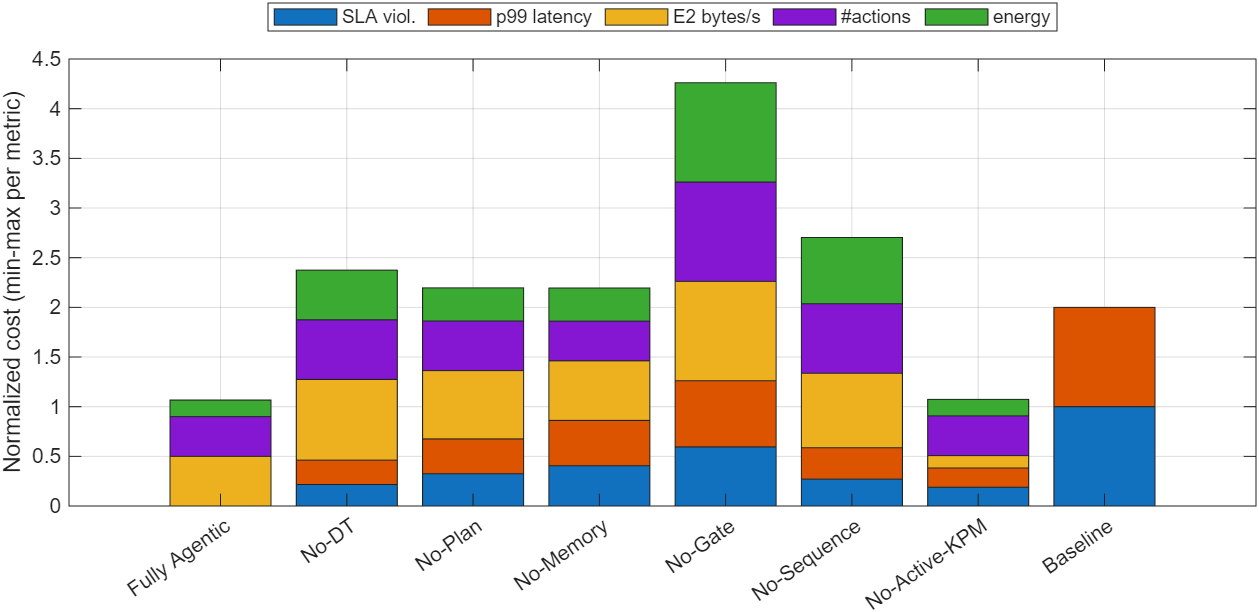} 
\caption{Stacked, min–max normalized KPIs for the full agentic controller, ablations, and the conventional baseline, illustrating the trade-off between QoS and operational cost.}
\label{fig:agentic-stacked-metrics}
\end{figure*}

\section{System Overhead and Scalability Analysis}

{The agentic primitives introduced in Section~IV add computational, memory, and signalling cost on top of the standard O-RAN control plane. This section discusses these costs qualitatively, examines how they scale with deployment size, and analyzes failure modes introduced by the LLM-assisted Non--RT layer.}

\subsection{Computational and Signalling Overhead}

{Overhead in the proposed framework is concentrated at the Non--RT RIC and is structured to remain bounded at lower layers. At the Non--RT layer, the LLM-based intent grounding and resource summarization run at minute-level cadence; each invocation processes a compact context (KPM/KQI summaries, slice-level aggregates) rather than raw telemetry, keeping prompt size on the order of a few thousand tokens. Inference latency of a state-of-the-art commercial LLM under this prompt budget is typically below one second per call, which is two orders of magnitude smaller than the Non--RT control cadence and therefore does not constrain end-to-end timing.
At the Near--RT RIC, agentic primitives operate without LLM calls. The skill sequencer, self-management gate, and short-term cache add only lightweight book-keeping per decision: schema checks on candidate skills, threshold comparisons against budget counters, and ring-buffer updates on KPM windows. Memory retrieval is implemented as a similarity lookup over a bounded episodic store and incurs sub-millisecond latency. As a result, the Near--RT decision loop remains comfortably within the 10\,ms--1\,s O-RAN budget.
Signalling overhead over the standardized interfaces is also bounded by design. A1 traffic increases only at the Non--RT cadence, since LLM-derived policies are dispatched as periodic JSON manifests. E2 overhead is dominated by active KPM sampling, which the agent throttles via the self-management gate when budget pressure is detected. The No-Active-KPM ablation in Section~VI quantifies this trade-off: turning off active sampling reduces E2 bytes per second but degrades SLA satisfaction, confirming that the agent uses the additional bandwidth purposefully rather than as overhead.}

\subsection{Scalability}

{Scalability concerns arise along three axes: number of user equipment (UE) devices, number of concurrent xApps, and size of the long-term memory.
The number of UEs primarily affects KPM volume and the size of the state summarized at the Non--RT layer. Because the agent operates on aggregated slice- and cell-level summaries rather than per-UE features, prompt size grows with the number of cells and slices rather than with UE count, scaling to typical urban deployments without requiring proportionally larger LLM contexts.
The number of concurrent xApps affects conflict resolution overhead at the Near--RT RIC. The skill sequencer detects conflicts based on scope intersection and predicted impact on shared KPMs, with cost scaling roughly quadratically in the number of overlapping skills. In practice, scope intersections are sparse since most skills act on disjoint cells or parameter groups, keeping the effective conflict graph small even with tens of concurrent xApps. When density grows further, the agent can partition the conflict graph by cell cluster, recovering near-linear scaling at the cost of mild loss of cross-cluster optimality.
Long-term memory growth is bounded by retention policies that retain only validated skill compositions and decision-level evidence summaries rather than raw telemetry. Retrieval is performed over compact embeddings, so query latency grows logarithmically with memory size and remains negligible relative to the Non--RT cadence. For very large deployments, the long-term memory can be sharded by domain (e.g., per slice type or per region), with cross-shard retrieval invoked only when local priors are insufficient.}

\subsection{Failure and Safety Analysis}
{Three failure modes are particularly relevant for an LLM-assisted O-RAN controller, and each is mitigated within the framework rather than externalized.
LLM API timeout or unavailability does not propagate to time-critical paths because the LLM is invoked only at the Non--RT cadence. On timeout, the Near--RT RIC continues executing the last validated skill plan under its local self-management gate, while the Non--RT agent falls back to the most recent vetted policy retrieved from long-term memory. The timeout event is logged to the evidence bus for operator review.
LLM hallucination of unsafe outputs (e.g., a slicing budget that violates declared resource ceilings or SLA floors) is intercepted by the goal-grounding module. All LLM outputs are produced as structured JSON and validated against hard schema constraints and feasibility bounds before being dispatched via A1. Outputs that fail validation are rejected and replaced with conservative defaults, with each rejection recorded as an audit event.
Digital twin synchronization drift can render twin-derived evidence stale during rapidly changing traffic conditions. The framework treats DT support as a best-effort source rather than a hard dependency: when twin-real divergence (measured via residual KPM mismatch) exceeds a threshold, the agent shrinks step size, falls back to live observations only, and tightens guard thresholds in the self-management gate. As conditions stabilize, twin-assisted planning resumes.
Together, these mechanisms ensure that overhead remains bounded, scalability degrades gracefully with deployment size, and component-level failures degrade to deterministic fallbacks instead of propagating into the live RAN.}

\section{Evaluation}
\subsection{Experimental Setup}
We consider a simulated O-RAN segment deployed over a 500m $\times$ 500m area with six O-RUs, each hosting two radio cells (RCs), for a total of 12 cells. The number of active UEs is 20 to emulate light to moderately loaded conditions. Propagation is modelled using a free-space path loss channel, and each RC transmits at 30 dBm over 5G New Radio (NR) bands N77 and N78. Mobility and cell association procedures are handled through a standard received-signal-strength-based handover criterion. This setting provides a controlled yet representative environment to benchmark different agentic control variants under increasing load and multi-cell, multi-band operation. In our simulations, each experiment consists of 50 independent episodes, each corresponding to a fixed network scenario with stochastic traffic and mobility realizations. Each episode lasts 600 seconds of simulated time.
{The agent's decision space is structured as follows. Observations include per-slice throughput/latency aggregates, per-cell physical resource block (PRB) utilization, transmit power, queue lengths, and handover event rates, sampled at 100 ms granularity from E2. Actions are skill invocations from a catalog including {power-cap, PRB-reallocate, handover (HO)-threshold-adjust, slice-scale-up, slice-scale-down} with bounded parameter ranges. Skill outputs are translated into E2/A1 control messages applied at the next decision tick, with outcomes observed in the subsequent KPM window and recorded as evidence.}

\subsection{Agentic Ablations}
\subsubsection{Fully Agentic}
The full Agentic controller implements planning over skills, sequencing, self-management gating, multi-horizon memory and retrieval, and active KPM sampling. All ablations are derived from this configuration by disabling specific primitives.
\subsubsection{No-plan}
The No-Plan variant disables multi-step planning and selects only a single next skill reactively from the current context. This isolates the benefit of explicit long-horizon planning over myopic decision making.

\subsubsection{No-memory}
The No-Memory variant removes episodic and long-term memory and retrieval, retaining only short-term KPM/KQI windows. It quantifies the contribution of case reuse and cross-episode knowledge transfer under changing traffic and mobility.
\subsubsection{No-gate}
In the No-Gate variant, skill sequences are executed without self-management checks on risk, uncertainty, budgets, or explanation consistency. Comparison with the full agent highlights the role of gating in reducing SLA violations and unstable behavior.
\subsubsection{No-sequence}

The No-Sequence variant restricts plans to a single skill, disabling composite multi-step interventions. This reveals the impact of skill sequencing compared to isolated knob adjustments.
\subsubsection{No-active-KPM}

The No-Active-KPM variant disables active measurement control and relies on a fixed set of passively collected KPMs. It evaluates the benefit of information-aware telemetry under an E2 budget.

\subsubsection{No-DT}
{The No-DT variant disables the digital twin as a planning and evidence source. The agent operates solely on live KPM/KQI observations without access to what-if simulation or twin-derived context for goal grounding. This isolates the contribution of DT-supported planning to slice admission accuracy and SLA stability, particularly under rare or unseen traffic conditions where historical memory alone is insufficient.}

\begin{table*}[t]
\centering
\caption{Impact of Non--RT LLM on Network Slicing Performance}
\label{tab:llm_slicing_impact}
\setlength{\tabcolsep}{4.5pt}
\renewcommand{\arraystretch}{1.15}
\begin{tabular}{c|ccc|ccc|ccc}
\hline
\multirow{2}{*}{Slice} 
& \multicolumn{3}{c|}{Admission Accuracy (\%) } 
& \multicolumn{3}{c|}{Resource Usage (\%) } 
& \multicolumn{3}{c}{99th-percentile (p99) Latency (ms) } \\
& No LLM & With LLM & Gain 
& No LLM & With LLM & Gain
& No LLM & With LLM & Gain \\
\hline
Enhanced Mobile Broadband (eMBB)  
& 88 & 93 & $\uparrow$5.7\%
& 68  & 61  & $\downarrow$10.3\%
& 18.0 & 15.0 & $\downarrow$16.7\% \\

Ultra-Reliable Low-Latency Communications (URLLC)
& 84 & 91 & $\uparrow$8.3\%
& 55  & 50  & $\downarrow$9.1\%
& 9.5 & 7.8 & $\downarrow$17.9\% \\

Massive Machine-Type Communications (mMTC)
& 86 & 90 & $\uparrow$4.7\%
& 42  & 39  & $\downarrow$7.1\%
& 35.0 & 30.0 & $\downarrow$14.3\% \\
\hline
\end{tabular}
\end{table*}

\subsection{LLM Benchmark}

\subsubsection{Conventional Baseline}
The learning baseline replaces the agentic controller with a standard deep RL xApp at the Near--RT RIC: a Deep Q-Learning (DQL) agent that maps a compact state vector to discrete actions (scale-up, scale-down, or no-op), trained offline on trace- or twin-driven episodes and frozen for online evaluation as a monolithic black-box controller without skill planning, structured memory, or self-management gating.

Figure~\ref{fig:agentic-stacked-metrics} summarizes five KPIs across all variants: \emph{SLA violation} (fraction of episodes in which any slice fails its throughput/latency target), \emph{$p99$ latency} (99th-percentile end-to-end packet latency per episode), \emph{E2 bytes per second} (average E2 message volume between Near-RT RIC and DU/CU), \emph{number of actions} (skill invocations per episode), and \emph{energy} (aggregate radio and compute consumption). For each metric $k$ and variant $v$, we interpret the metric as a cost (larger is worse) and apply a per-metric min--max normalization:
\begin{equation}
\tilde{m}_{v,k} = \frac{m_{v,k} - \min_{v'} m_{v',k}}{\max_{v'} m_{v',k} - \min_{v'} m_{v',k}},
\end{equation}
so that the best variant on metric $k$ has $\tilde{m}_{v,k}=0$ and the worst has $\tilde{m}_{v,k}=1$. Each bar stacks $\tilde{m}_{v,k}$ across the five metrics, making the relative contribution of each cost dimension visually comparable.
{The mechanistic interpretation of Fig.~\ref{fig:agentic-stacked-metrics} follows from which primitive each variant disables. No-Gate produces the largest increase in energy, E2 overhead, and action count, because skills execute without budget or risk pre-checks; large-amplitude actions trigger side-effects that subsequent commits must compensate, leading to action cascades. No-Plan modestly raises SLA violations by decomposing multi-step interventions (e.g., power capping before PRB reallocation in interference-limited regimes) into uncoordinated single steps, increasing convergence time. No-Memory raises tail latency because the agent rediscovers parameter combinations online instead of retrieving validated compositions. No-Sequence cannot apply composite remedies, leaving residual SLA pressure as elevated $p99$ latency. No-Active-KPM trades reduced E2 bytes for SLA risk: passive telemetry misses transient indicators and forces corrective follow-ups. No-DT loses what-if validation; its impact is moderate on average but pronounced under rare conditions such as flash crowds{, consistent with the DT's role as an online buffer that pre-stages responses to extreme scenarios (Section~III-E)}. The DQL baseline exhibits the opposite trade-off: low operational overhead but high SLA and tail-latency cost, reflecting scalar-reward encoding of multi-objective constraints and the absence of explicit gating against distribution shifts.}

\subsubsection{LLM Coordination at Non-RT RIC}
To assess the system-level value of introducing an LLM at the Non--RT RIC, we compare agentic control with and without LLM-based slicing coordination. Due to the strict timing requirements of O-RAN, the LLM operates only at minute-level intervals to summarize Near--RT reinforcement-learning behaviors, aggregate slice-level performance indicators, and refine slicing budgets and controller activation decisions. {The Near-RT closed-loop xApps remain unchanged.}{This setup directly addresses the organizational-versus-content question: the RL content optimizers are held fixed and the agentic LLM layer is added only as an organizational coordinator on top, so any improvement is attributable to organization rather than to a stronger underlying optimizer. Table~II is therefore evidence of \emph{composition}, not of one paradigm displacing the other.} We employ a state-of-the-art commercial LLM (GPT-5.2-class \cite{openai2025gpt52}) as the underlying reasoning engine, with a context window of approximately 8k tokens and generation temperature set to 0.2 to favor deterministic JSON outputs. The LLM produces structured JSON manifests containing slice-level resource budgets and controller activation flags, which are schema-validated by the goal-grounding module against hard resource ceilings and SLA floors before being dispatched via A1 (a representative example is provided in Appendix~B). Outputs that violate declared constraints are rejected and replaced with conservative defaults retrieved from long-term memory, with each rejection logged to the evidence bus.

Table~\ref{tab:llm_slicing_impact} shows that Non--RT LLM coordination consistently improves network slicing outcomes across service classes, increasing slice admission accuracy, reducing overall resource consumption, and lowering tail latency by enabling more informed cross-slice allocation decisions. These gains demonstrate that LLMs can enhance resource efficiency and slicing precision when confined to non-real-time layers, without violating O-RAN control-loop constraints or introducing LLM-based reasoning into time-critical RAN paths.

\section{Conclusion and Future Directions}
 This article proposed an Agentic AI-RAN perspective, in which coordinated, goal-driven agents are embedded across the SMO/Non-RT RIC, Near-RT RIC, and O-CU/O-DU to operate over O-RAN skills with explicit timing and budget awareness. {The design combines structured planning, skill-based control, multi-timescale memory and evidence, and self-management gating, and remains interoperable with existing O-RAN specifications.  existing O-RAN specifications. {Crucially, the agentic layer is positioned as an organizational coordinator over---rather than a replacement for---RL/MARL content optimizers: the Non-RT LLM experiment improves slicing outcomes while leaving the Near-RT RL xApps unchanged, indicating that the two abstraction levels compose rather than compete.}}

{Multi-cell simulations show that a fully agentic controller improves SLA satisfaction and stability over ablated variants and deep RL baselines, while making trade-offs in control overhead and energy transparent and auditable. Future work will integrate richer RL/MARL components within the agentic framework, evaluate the design in higher-fidelity digital-twin environments, and explore standardized hooks—skill catalogues, evidence buses, and policy-as-code guardrails—for operational deployment.}

\section*{Acknowledgment}
This work was supported in part by “Funded through the Open Challenge Fund call, organised by the Datacom Industry Association AISBL”. The authors acknowledge the funding and support provided by the EPSRC UK-India Future Networks Initiative (UKI-FNI). Details of the project can be found at https://www.ukifni.org/.  This work was supported in part by the EPSRC and DSIT through the Communications Hub for Empowering Distributed Cloud Computing Applications and Research (CHEDDAR) [grant numbers EP/X040518/1 and EP/Y037421/1].

\bibliographystyle{IEEEtran}
\bibliography{MyRef.bib}

@article{he2025digital,
  title={Digital Twin-Enhanced Reinforcement Learning for Intelligent xApps Management in {O-RAN} Systems},
  author={He, Zhizhou and Al-Tahmeesschi, Ahmed and Foh, Chuan Heng and Ahmadi, Hamed and Shojafar, Mohammad},
  journal={IEEE Internet Things Mag.},
  year={2025},
  publisher={IEEE}
}

@inproceedings{he2025heterogeneous,
  title={Heterogeneous-Agent {PPO} {RL} for xApps Coordination in Digital Twin Enabled {O-RAN}},
  author={He, Zhizhou and Luo, Yang and Shojafar, Mohammad and Mi, De},
  booktitle={Proc. IEEE/CIC Int. Conf. Commun. China (ICCC)},
  pages={1--6},
  year={2025},
  organization={IEEE}
}

@article{chatzimiltis2025AI,
  title={{AI}-on-RAN for cyber defense: An X{AI}-[{LLM}] framework for interpretable anomaly detection},
  author={Chatzimiltis, Sotiris and Shojafar, Mohammad and Mashhadi, Mahdi Boloursaz and Tafazolli, Rahim},
  journal={IEEE Transactions on Network Science and Engineering},
  volume={13},
  pages={3301--3319},
  year={2025},
  publisher={IEEE}
}

@article{dev2025advancedAI,
  title={Advanced architectures integrated with agentic {AI} for next-generation wireless networks},
  author={Dev, Kapal and Khowaja, Sunder Ali and Zeydan, Engin and Singh, Keshav and Debbah, Merouane},
  journal={IEEE Communications Standards Magazine},
  year={2025},
  publisher={IEEE}
}

@article{elkael2025agentran,
  title={{AgentRAN}: An Agentic {AI} Architecture for Autonomous Control of Open {6G} Networks},
  author={Elkael, Maxime and D'Oro, Salvatore and Bonati, Leonardo and Polese, Michele and Lee, Yunseong and Furueda, Koichiro and Melodia, Tommaso},
  journal={arXiv preprint arXiv:2508.17778},
  year={2025}
}

@ARTICLE{chatzimiltis2025agentic,
  author={Chatzimiltis, Sotiris and Mashhadi, Mahdi Boloursaz and Shojafar, Mohammad and Debbah, Merouane and Tafazolli, Rahim},
  journal={IEEE Communications Standards Magazine}, 
  title={Agentic {AI} for {6G}: A New Paradigm for Autonomous {RAN} Security Compliance}, 
  year={2026},
  volume={},
  number={},
  pages={1-9},
  doi={10.1109/MCOMSTD.2026.3681685}}

@article{alam2025comprehensive,
  title={A comprehensive tutorial and survey of {O-RAN}: Exploring slicing-aware architecture, deployment options, use cases, and challenges},
  author={Alam, Khurshid and Habibi, Mohammad Asif and Tammen, Matthias and Krummacker, Dennis and Saad, Walid and Di Renzo, Marco and Melodia, Tommaso and Costa-P{\'e}rez, Xavier and Debbah, M{\'e}rouane and Dutta, Ashutosh and others},
  journal={IEEE Commun. Surv. Tutor.},
  year={2025},
  publisher={IEEE}
}

@article{ngo2024ran,
  title={{RAN} Intelligent Controller ({RIC}): From open-source implementation to real-world validation},
  author={Ngo, Mao V and Yoo, Hyun-Min and Pua, Yong-Hao and Le, Thanh-Long and Liang, Xian-Loong and Chen, Binbin and Hong, Een-Kee and Quek, Tony QS and others},
  journal={ICT Express},
  volume={10},
  number={3},
  pages={680--691},
  year={2024},
  publisher={Elsevier}
}

@article{sroka2024policy,
  title={Policy-based traffic steering and load balancing in {O-RAN}-based vehicle-to-network communications},
  author={Sroka, Pawe{\l} and Ku{\l}acz, {\L}ukasz and Janji, Salim and Dryja{\'n}ski, Marcin and Kliks, Adrian},
  journal={IEEE Trans. Veh. Technol.},
  volume={73},
  number={7},
  pages={9356--9369},
  year={2024},
  publisher={IEEE}
}

@article{polese2023understanding,
  title={Understanding {O-RAN}: Architecture, interfaces, algorithms, security, and research challenges},
  author={Polese, Michele and Bonati, Leonardo and D’oro, Salvatore and Basagni, Stefano and Melodia, Tommaso},
  journal={IEEE Commun. Surv. Tutor.},
  volume={25},
  number={2},
  pages={1376--1411},
  year={2023},
  publisher={IEEE}
}

@article{polese2023empowering,
  title={Empowering the {6G} cellular architecture with open {RAN}},
  author={Polese, Michele and Dohler, Mischa and Dressler, Falko and Erol-Kantarci, Melike and Jana, Rittwik and Knopp, Raymond and Melodia, Tommaso},
  journal={IEEE J. Sel. Areas Commun.},
  volume={42},
  number={2},
  pages={245--262},
  year={2023},
  publisher={IEEE}
}

@article{brik2024explainable,
  title={Explainable {AI} in {6G} {O-RAN}: A tutorial and survey on architecture, use cases, challenges, and future research},
  author={Brik, Bouziane and Chergui, Hatim and Zanzi, Lanfranco and Devoti, Francesco and Ksentini, Adlen and Siddiqui, Muhammad Shuaib and Costa-P{\`e}rez, Xavier and Verikoukis, Christos},
  journal={IEEE Commu. Surv. Tutor.},
  year={2024},
  publisher={IEEE}
}

@article{xu2024large,
  title={When large language model agents meet 6{G} networks: Perception, grounding, and alignment},
  author={Xu, Minrui and Niyato, Dusit and Kang, Jiawen and Xiong, Zehui and Mao, Shiwen and Han, Zhu and Kim, Dong In and Letaief, Khaled B},
  journal={IEEE Wireless Communications},
  volume={31},
  number={6},
  pages={63--71},
  year={2024},
  publisher={IEEE}
}

@misc{openai2025gpt52,
  author       = {{OpenAI}},
  title        = {Introducing {GPT}-5.2},
  year         = {2025},
  month        = dec,
  howpublished = {\url{https://openai.com/index/introducing-gpt-5-2/}},
  note         = {Accessed: 2026-05-11}
}

\end{document}